\documentclass[12pt]{article}
\usepackage[margin=1in]{geometry}
\usepackage{float} 
\usepackage{caption}
\usepackage{amsmath}
\usepackage{graphicx}
\usepackage{hyperref}
\usepackage{caption}
\usepackage{amsfonts}
\usepackage{float}
\usepackage{setspace}
\usepackage{indentfirst}
\usepackage{titlesec}
\usepackage{algorithm}
\usepackage{algorithmic}
\usepackage{subcaption} 
\usepackage{authblk}

\title{\textbf{Image Classification Using Singular Value Decomposition and Optimization}}
\author[1*]{Isabela M. Yepes}
\author[1*]{Manasvi Goyal}
\affil[1]{Harvard University, Cambridge, MA, 02138, USA}

\date{}

\begin{document}

\singlespacing
\maketitle
\vspace{-33pt}
\vspace{2pt}

\def\thefootnote{*}\footnotetext{These authors contributed equally to this work.}

\begin{abstract} 
This study investigates the applicability of Singular Value Decomposition for the image classification of specific breeds of cats and dogs using fur color as the primary identifying feature. Sequential Quadratic Programming (SQP) is employed to construct optimally weighted templates. The proposed method achieves 69\% accuracy using the Frobenius norm at rank 10. The results partially validate the assumption that dominant features, such as fur color, can be effectively captured through low-rank approximations. However, the accuracy suggests that additional features or methods may be required for more robust classification, highlighting the trade-off between simplicity and performance in resource-constrained environments.

\end{abstract}

\vspace{-8pt}

\section{Introduction}

Traditional classification methods often rely on machine learning or deep learning models, which require significant computational resources and large annotated datasets. These methods, while effective, can be challenging to deploy in resource-constrained environments. Singular Value Decomposition (SVD) is known for its ability to reduce the dimensionality of images. It captures the essential features of images through low-rank approximations \cite{heath2005scientific}. While SVD has been extensively applied in areas such as image compression and noise reduction, its application to direct image classification remains relatively unexplored.

This study proposes using SVD in conjunction with optimization techniques as an alternative to conventional machine learning methods for image classification in resource-constrained settings. The proposed method involves first constructing a representative template for each image class using a training set of images. Two methods for constructing templates are evaluated, one uses an optimally weighted approach with Sequential Quadratic Programming (SQP), and another method uses uniform weights. Then identifying the optimal rank and norm across the whole training set such that the overall model accuracy is maximized. For the test set of images, the prediction is computed by comparing the optimal low-rank approximation to the training set template using the optimal norm.

This method provides an alternative to machine learning models and instead focuses on taking advantage of the properties of matrix decompositions and error minimization for classification. The feasibility of the method is evaluated by examining reconstruction errors computed using various ranks and norms to identify optimal parameters for classification.

\section{Data}

The dataset used in this study is the \textit{23 Pet Breed Image Classification} dataset publicly available on Kaggle \cite{kaggle2023pets}. The dataset contains images of dogs and cats. The study focuses on two specific classes, Persian cats and Boxer dogs, each comprising 170 images. All images in the dataset are in \texttt{jpg} or \texttt{jpeg} format.

\section{Methodology}

This proposed method will be applied to a binary image class dataset of Boxer dogs and Persian cats. The hypothesis for applying this model to this dataset relies on the assumption that the fur color is the visually distinctive class identifier with Persian cats typically having white fur, and Boxer dogs generally having a darker fur color. Further, the hypothesis is that the fur color is a feature that can be effectively captured at a low rank approximation. The methodology consists of three main components: image preprocessing, template creation, and image classification. These steps are explained in the following sections.

\subsection{Image Preprocessing}

The first step in the pipeline is preprocessing the raw images to ensure consistency in size and format. Each image is represented as a matrix \( A \in \mathbb{R}^{m \times n} \), where \( m = n = 256 \) after resizing. The images are converted from RGB to grayscale using the \verb"rgb2gray" method of the \verb"color" module in the \verb"skimage" library in Python. All images are then resized to \( 64 \times 64 \) pixels. The resizing step ensures that all images have uniform dimensions. Additionally, the pixel intensities are normalized to the range \([0, 1]\), which improves numerical stability during SVD computations.

\subsection{Template Creation}

Templates serve as representative summaries of each image class. These templates encapsulate the dominant features and patterns shared among training images of the same class. By reducing the variability present in individual images, templates aim to improve the robustness of the classification process, particularly when dealing with noisy or high-dimensional data. Representative templates for each class are constructed using two approaches: the \textit{uniformly weighted} approach and the \textit{optimally weighted} approach.

\subsubsection{Uniformly Weighted Template}
First, the templates for each class \( C \) are computed as the average of all training images within a class. Given a set of training images \( A_1, A_2, \dots, A_N \), the template is computed as shown in Equation \ref{eq:avg}. The uniformly weighted template \( \overline{T}_C \) captures the dominant features shared by all images in the class, such as general shape and structure. By averaging multiple images, this approach inherently reduces noise and removes unimportant features, creating a basic representation of the class.
\[
\overline{T}_C = \frac{1}{N} \sum_{i=1}^N A_i \tag{1} \label{eq:avg}
\]

\subsubsection{Optimally Weighted Template}
To further refine the templates, a weighted optimization approach is employed. Each image \( A_i \) is vectorized into a column vector \( \mathbf{a}_i \in \mathbb{R}^{m \times n} \). The template for a given class is computed as a weighted average as expressed in Equation \ref{eq:weighted}.
\[
T_C = \sum_{i=1}^N w_i \mathbf{a}_i  \tag{2} \label{eq:weighted}
\]

The weights \( w_i \) are optimized to minimize the reconstruction error as:
\[
E(\mathbf{w}) = \sum_{i=1}^N \| \mathbf{a}_i - T_C \|^2  \tag{3} \label{eq:error}
\]

The optimization problem is solved using the \texttt{scipy.optimize.minimize} function with the Sequential Least Squares Programming (SLSQP) method \cite{scipy}. SLSQP is a specific implementation of SQP to solve nonlinear optimization efficiently \cite{kraft}. While SQP handles both nonlinear and linear constraints by iteratively linearizing them within quadratic programming subproblems, SLSQP focuses specifically on problems with linear constraints and is particularly suited for least squares formulations \cite{neos, cornell}. SLSQP is particularly well-suited for this task as it efficiently handles the linear constraints imposed on the weights \( \mathbf{w} \) \cite{kraft}. It is used to refine the template \( T \) by assigning higher weights \( w_i \) to training images \( \mathbf{a}_i \) that are more representative of the class. The method iteratively approximates the objective function in Equation \ref{eq:error}, with a quadratic model, and the constraints are approximated with linear models. This ensures that images closely aligned with the dominant features of the class contribute more to the template, while outliers and noisy images have minimal influence due to their smaller weights.

The method enforces two constraints described as follows:
\begin{enumerate}
    \item An \textbf{equality constraint}: \( \sum_{i=1}^N w_i = 1 \), which ensures the weights sum to 1.
    \item An \textbf{inequality constraint}: \( w_i \geq 0 \), implemented as \( w_i \geq \epsilon \) (a small positive value) to account for numerical precision and prevent negative weights.
\end{enumerate}

The optimization begins with an initial uniform guess for the weights, \( w_i = \frac{1}{N} \), ensuring the starting point satisfies the constraints. At each iteration, SLSQP updates the weights \( \mathbf{w} \) to minimize the reconstruction error \( E(\mathbf{w}) \), while maintaining feasibility with respect to the constraints. The process terminates when the change in \( \mathbf{w} \) between iterations or the improvement in \( E(\mathbf{w}) \) falls below a predefined tolerance.

The \textit{optimally} weighted template is expected to yield equal or improved image classification performance based on its objective function, compared to the \textit{uniformly} weighted template which serves as a baseline. This allows the template to better capture the dominant features of each class while balancing computational simplicity and precision.

\subsection{Image Classification Algorithm}

For the classification of a test image, the image is first preprocessed by converting it to grayscale and resizing it to a uniform size. Subsequently, the image is decomposed using SVD for a best rank \( k\) which is computed after a detailed analysis of reconstruction errors on the training set. This decomposition splits the test image into orthogonal components, enabling rank-\( k \) approximations to be constructed by retaining only the top \( k \) singular values and their corresponding singular vectors. The classification procedure is described in Algorithm \ref{alg:classification}.

For each class \( C \in \{\text{boxer}, \text{persian}\} \), the reconstruction error \(\text{Error}_{C, k}\) is computed between the rank-\( k\) approximation of the test image and full-rank template \( T_C \) of the class.

The norm \( \| \cdot \|_{\text{Norm}} \) is selected by comparing the classification accuracies of different norms on the training set. The class with the smallest reconstruction error is chosen as the predicted class for the given test image using the best rank and norm computed from the training set. 

\begin{algorithm}[H]
\caption{Classification of Test Image}
\label{alg:classification}
\textbf{Input:} Test image \( A_{\text{test}} \), Templates \( T_{\text{boxer}}, T_{\text{persian}} \).\\
\textbf{Output:} Predicted class \( \hat{C} \).
\begin{algorithmic}[1]
    \FOR{\( C \in \{\text{boxer, persian}\} \)}
        \STATE \( A_{\text{test}, k} \gets U_k \Sigma_k V_k^\top \)
        \STATE \( \text{Error}_{C} \gets \| A_{\text{test}} - T_{C} \|_{\text{Norm}} \)    
    \ENDFOR
    \IF{\( \text{Error}_{\text{boxer}} < \text{Error}_{\text{persian}} \)}
        \STATE \( \hat{C} \gets \text{"Boxer Dog"} \)
    \ELSE
        \STATE \( \hat{C} \gets \text{"Persian Cat"} \)
    \ENDIF
    \RETURN \( \hat{C} \)
\end{algorithmic}
\end{algorithm}

\section{Results}

\begin{figure}
    \centering
    \includegraphics[width=0.5\linewidth]{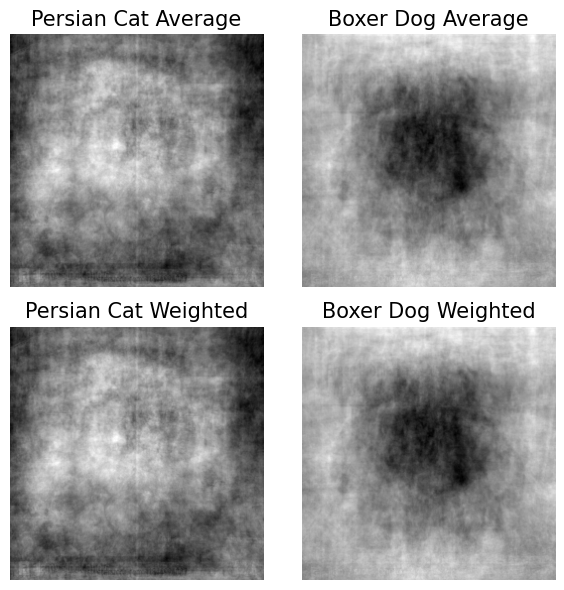}
    \caption{Persian Cat and Boxer Dog templates for uniform and optimal weights.}
    \label{fig:templates}
\end{figure}

The impact of different matrix norms and rank approximations on the reconstruction error and predictions is investigated to identify optimal parameters for image classification. First, the two proposed methods for computing a template from the training set are compared. The comparison of the uniformly and optimally weighted templates, denoted as average and weighted respectively, is shown in Figure \ref{fig:templates}. The sum of the difference in weights for the two methods applied to the Boxer dog template is \(1.82 \times 10^{-17} \), while it is \( -1.73 \times 10^{-18} \) for the Persian cat template. These differences are on the order of machine precision, specifically the IEEE 754 double-precision floating-point format (\texttt{binary64}), which has a precision of approximately \( 1.11\times 10^{-16} \) and thus it is negligible \cite{heath2005scientific}. Both the visual comparison and the difference in weights support the conclusion that the templates are equivalent. The rest of the analysis continues with the use of the optimally weighted template.


\subsection{Rank Selection}

Rank selection is important for determining the level of approximation used for test images which directly affects the reconstruction error and classification accuracy. Lower ranks capture coarse features, while higher ranks include finer details. Optimal rank selection ensures the best trade-off between these factors balancing computational efficiency and classification accuracy.

\begin{figure}
    \centering
    \begin{subfigure}{0.45\textwidth}
        \includegraphics[width=1.06\linewidth]{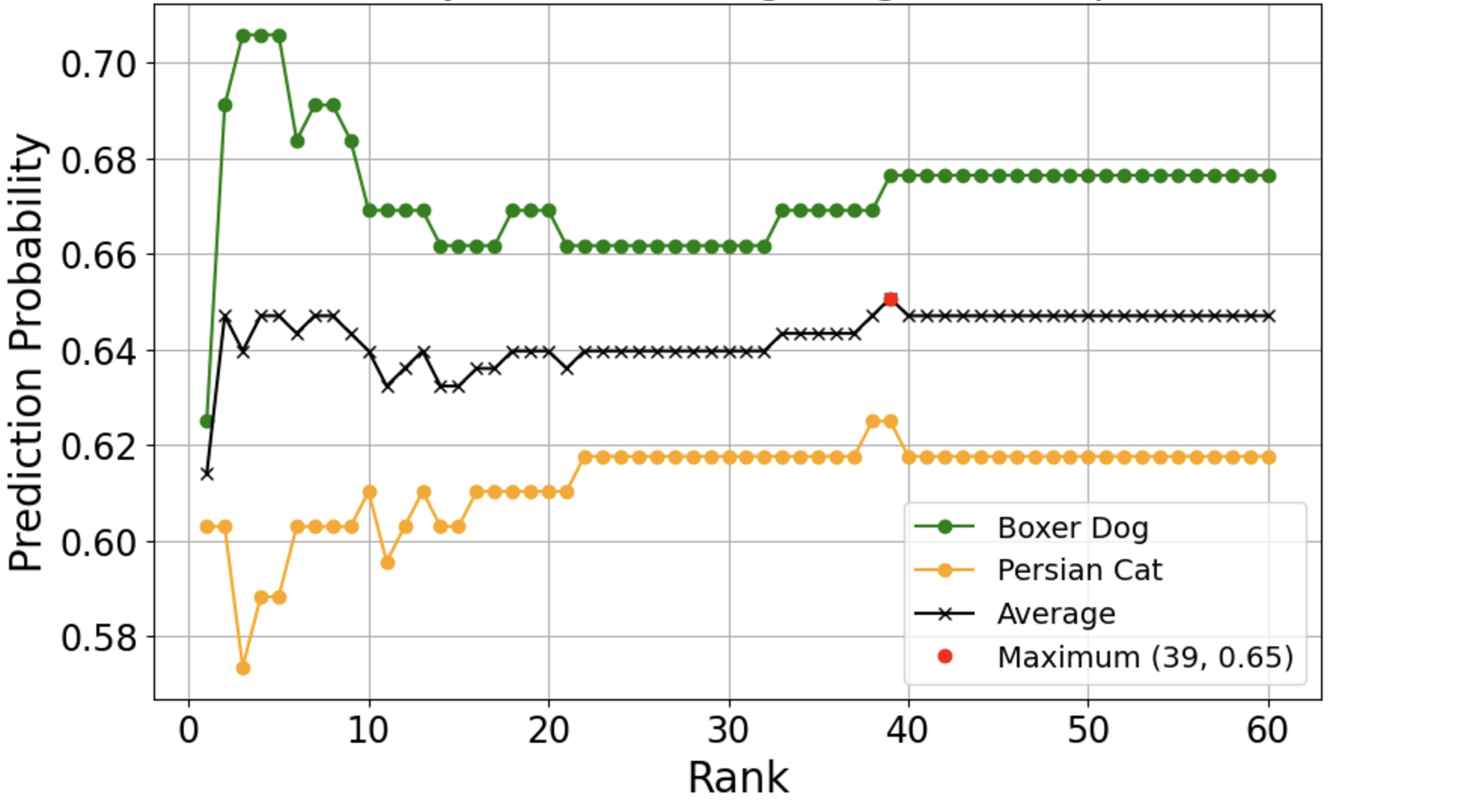} 
        \caption{1 norm}
        \label{fig:norm1}
    \end{subfigure}
    \hspace{5pt}
    \begin{subfigure}{0.45\textwidth}
        \includegraphics[width=1.05\linewidth]{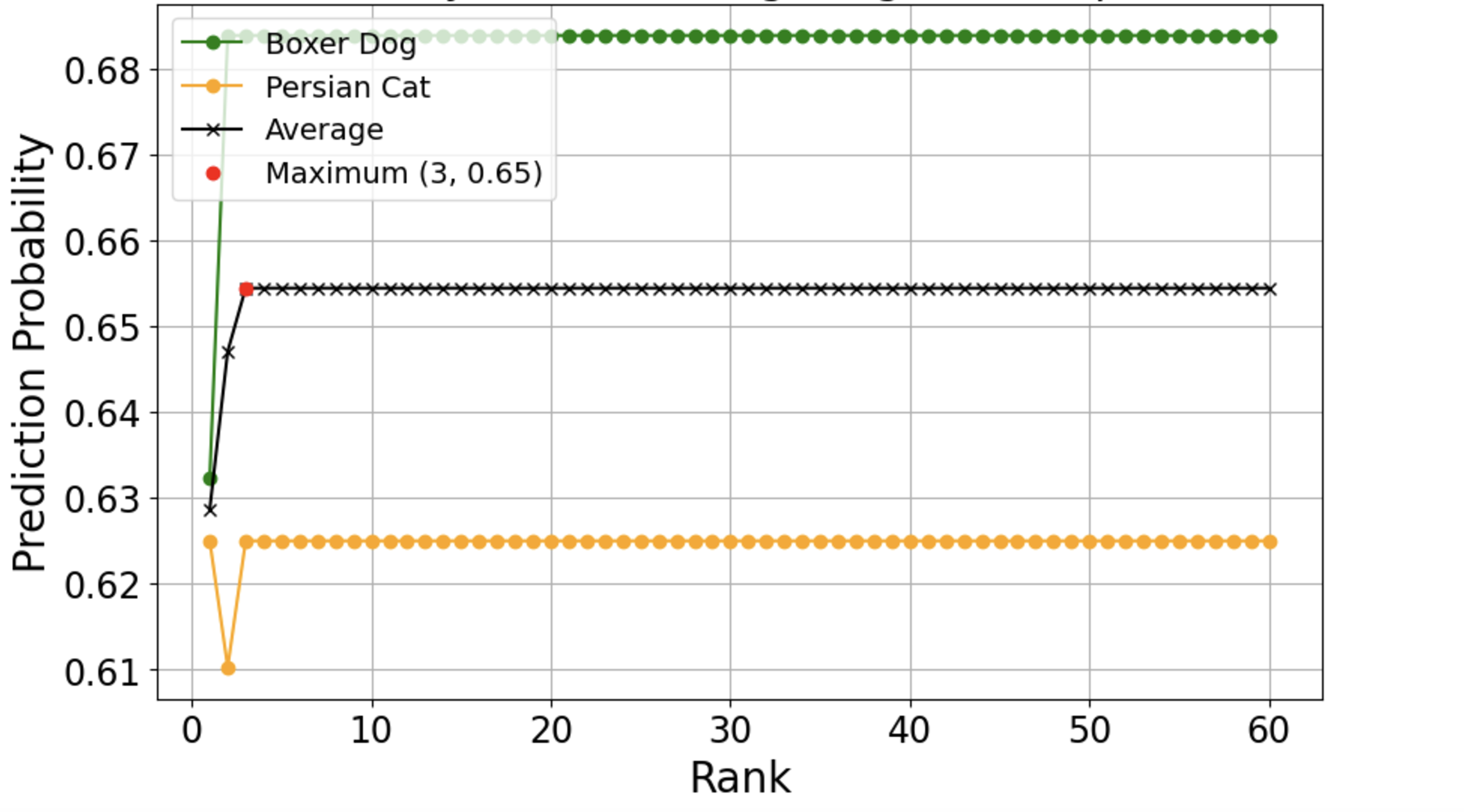} 
        \caption{2 norm}
        \label{fig:norm2}
    \end{subfigure}
    
    \vspace{0.5cm} 
    
    \begin{subfigure}{0.45\textwidth}
        \includegraphics[width=1.06\linewidth]{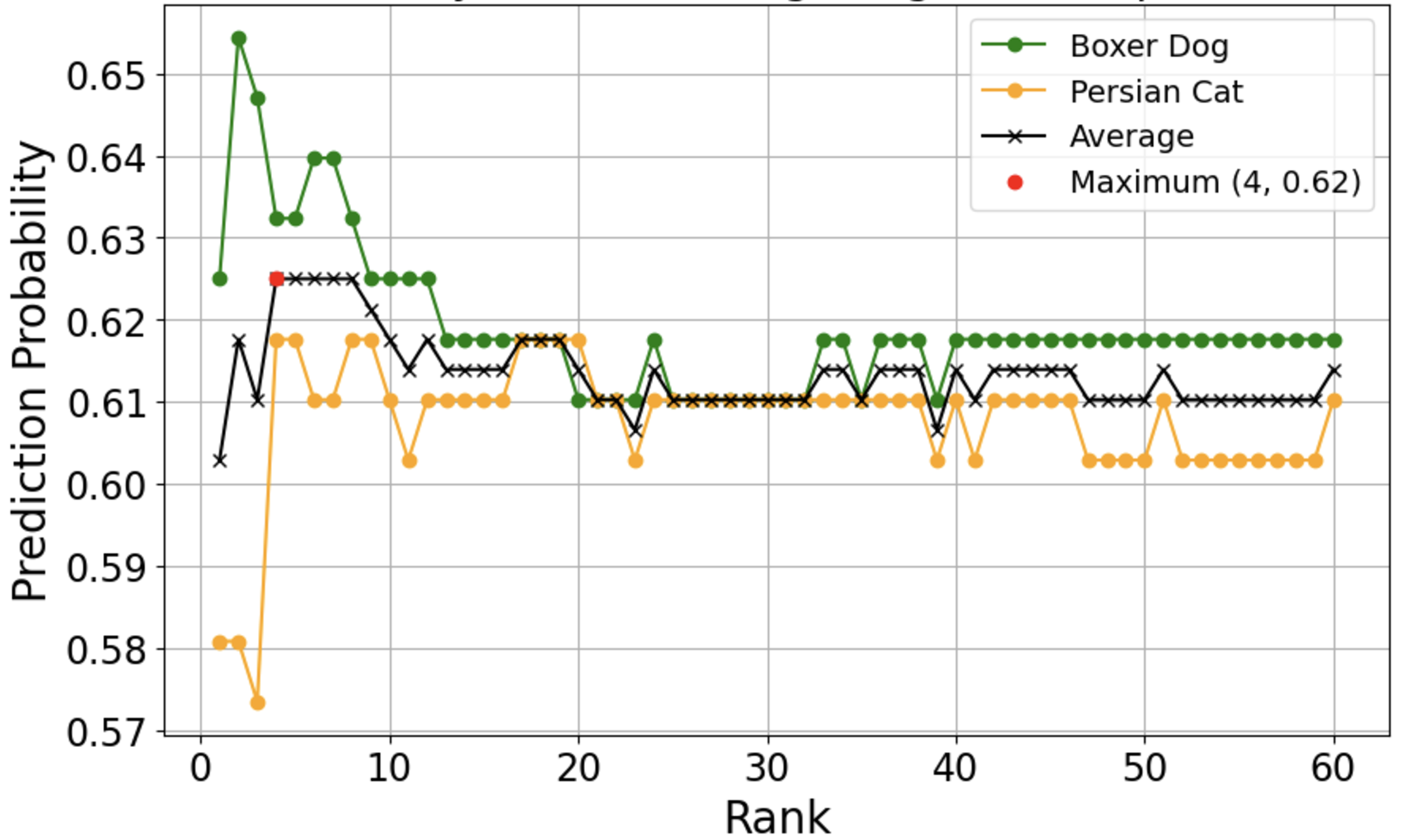} 
        \caption{\(\infty \) norm}
        \label{fig:norm_inf}
    \end{subfigure}
    \hspace{5pt}
    \begin{subfigure}{0.45\textwidth}
        \includegraphics[width=1.05\linewidth]{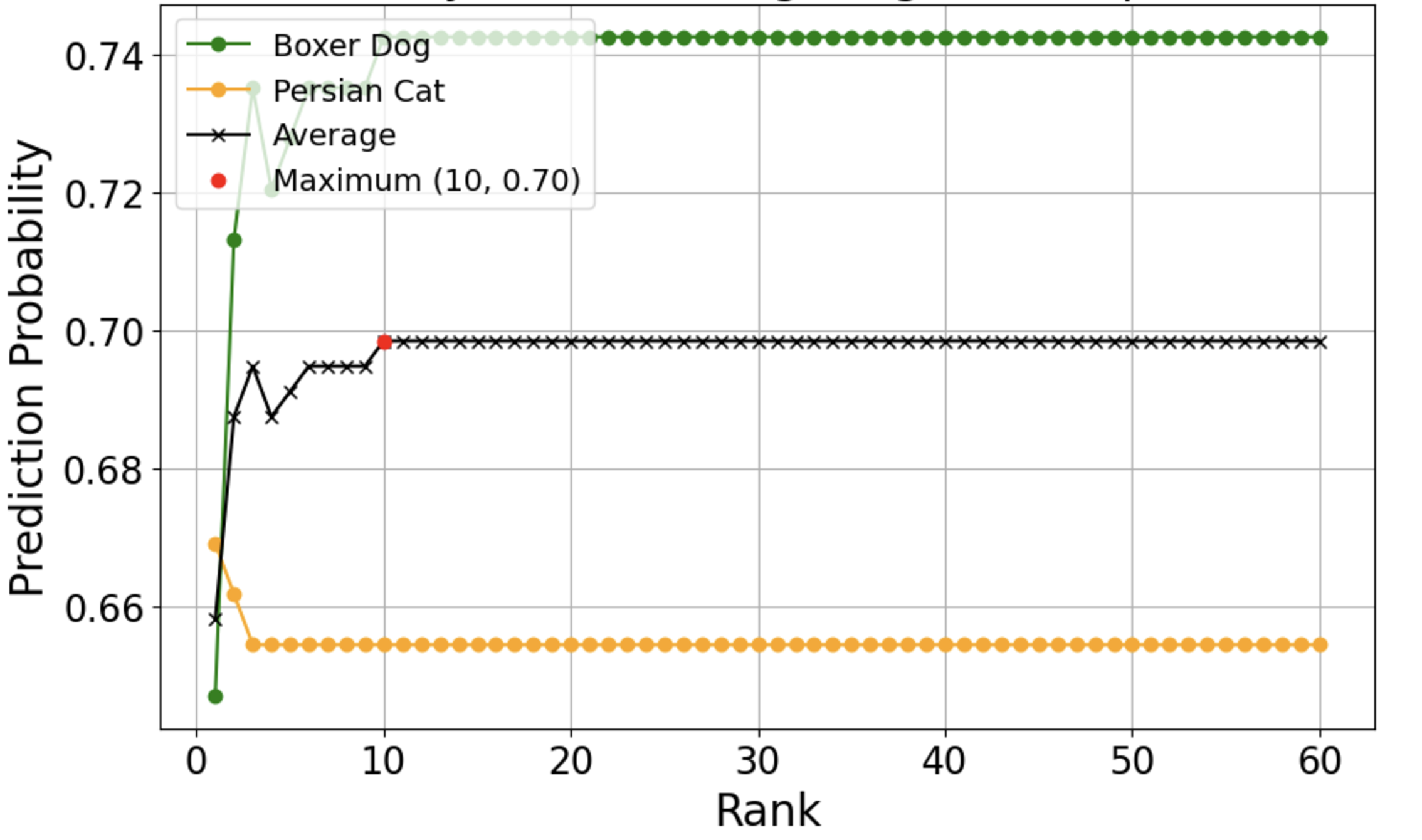} 
        \caption{Frobenius norm}
        \label{fig:norm_fro}
    \end{subfigure}
    
    \caption{Comparison of norms: 1 norm, 2 norm, \(\infty\) norm, and Frobenius norm across different ranks for the training set.}
    \label{fig:norms_grid}
\end{figure}

\begin{figure}
    \centering
    \includegraphics[width=1\linewidth]{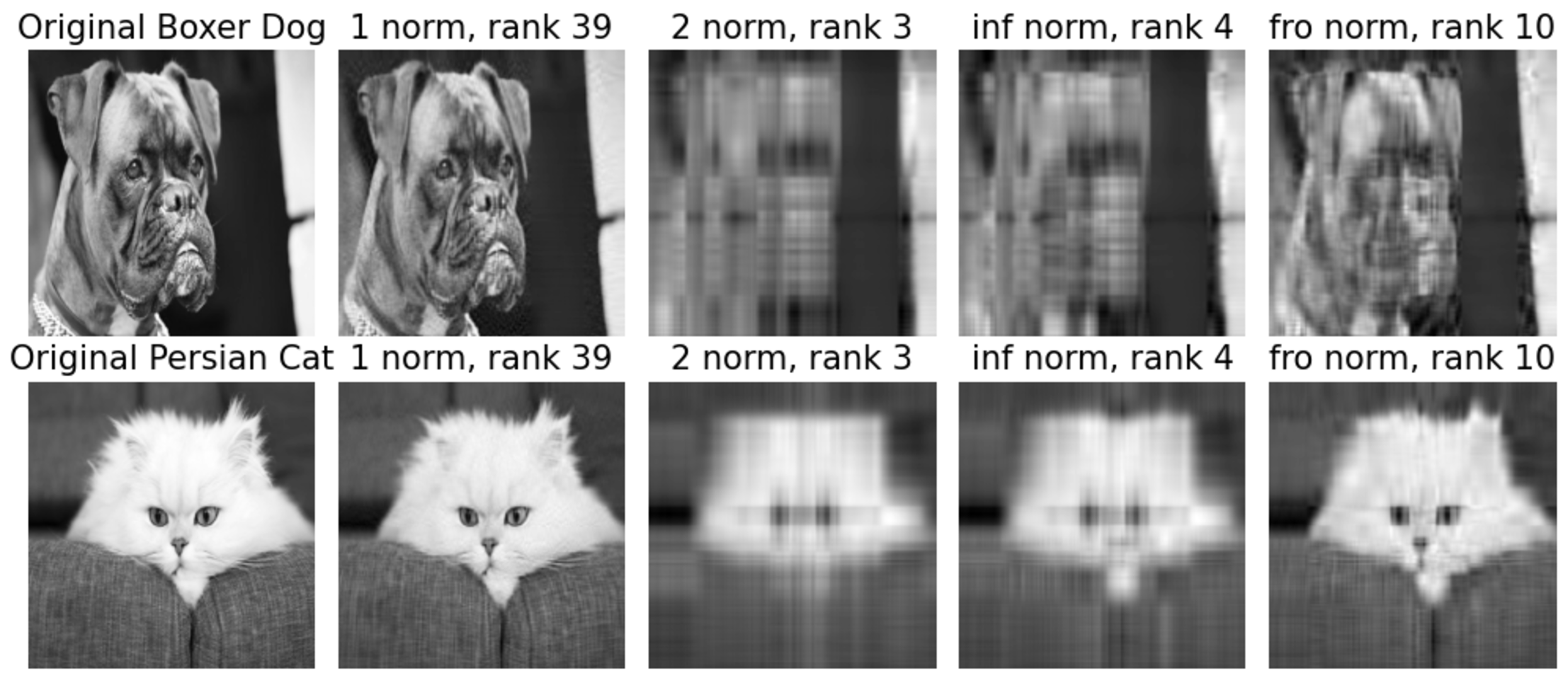}
    \caption{Low-rank representations of a test Persian cat and a test Boxer dog for each norm at the best rank.}
    \label{fig:test_all_norm}
\end{figure}

The errors in the training set are computed for different ranks of SVD approximations. Figure \ref{fig:norms_grid} shows that for all norms after a certain rank the prediction probabilities for each class converge. Further, the black line represents the average prediction probability of the two classes. The maximum of these average prediction probabilities is used to select the optimal rank for each norm. The optimal ranks for the norms: 1, 2, infinity, and frobenius, respectively are: 39, 3, 4, and 10. Test images for Persian cat and Boxer dog are visually represented at the optimal rank for each norm in Figure \ref{fig:test_all_norm}.

\subsection{Norm Selection}

Norm selection is crucial for evaluating reconstruction error and optimizing classification. Each norm captures distinct characteristics of the matrix representation of an image: the \( 1 \)-norm highlights the column with the largest total intensity, capturing dominant vertical structures; the \( 2 \)-norm reflects the largest singular value, emphasizing the most prominent feature of the image; the Frobenius norm measures the overall energy or texture of the image, treating all pixels equally; and the \( \infty \)-norm focuses on the row with the largest total intensity, capturing dominant horizontal structures. These differences influence how errors are measured and, consequently, classification accuracy.

\begin{figure}
    \centering
    \includegraphics[width=0.75\linewidth]{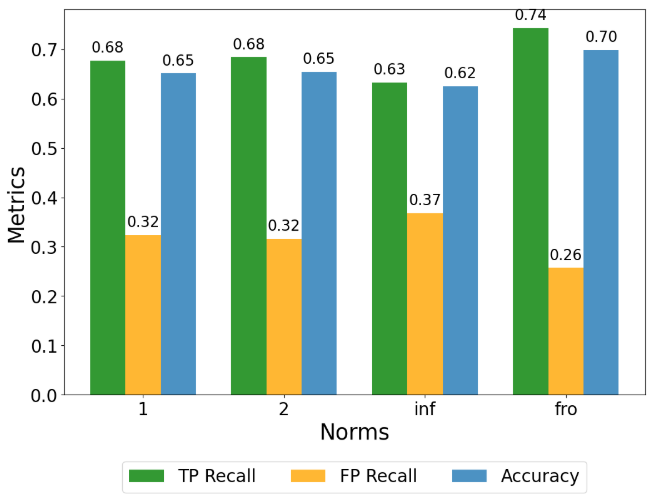}
    \caption{Classification metrics: TP Recall, FP Recall, and Accuracy across norms at respective optimal ranks for the training set.}
    \label{fig:bar_plt_norms}
\end{figure}

Figure \ref{fig:norms_grid} provides a comparison of the matrix norms in the training set. Note that the average prediction probability metric at the optimal rank is equivalent to the accuracy metric in Figure \ref{fig:bar_plt_norms}. Figure \ref{fig:bar_plt_norms} compares image classification performance for each norm evaluated at its optimal rank. It uses different metrics including TP Recall, FP Recall, and accuracy. Since the Frobenius norm achieves the highest accuracy, the rest of the analysis continues with the use of the Frobenius norm.

\subsection{Classification Results}
\begin{figure}
    \centering
    \includegraphics[width=0.75\linewidth]{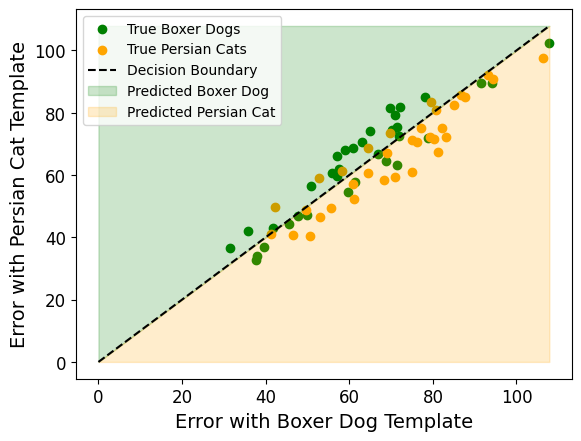}
    \caption{Test set classification of Boxer dogs vs. Persian cats, frobenius norm, rank = 10.}
    \label{fig:scatter}
\end{figure}

The classification results of the test set of images are computed with the optimal rank, the optimal norm, and class templates from the training dataset. In this case, the optimal norm is the Frobenius norm and the optimal rank is 10. The classification scatter plot for the test data set is shown in Figure \ref{fig:scatter} . Visually, the orange dots, representing true Persian cat images, are predominantly clustered within the orange-predicted Persian cat region, while the green dots, representing true Boxer dog images, are primarily distributed within the green-predicted Boxer dog region.

\begin{figure}[H]
    \centering
    \includegraphics[width=0.75\linewidth]{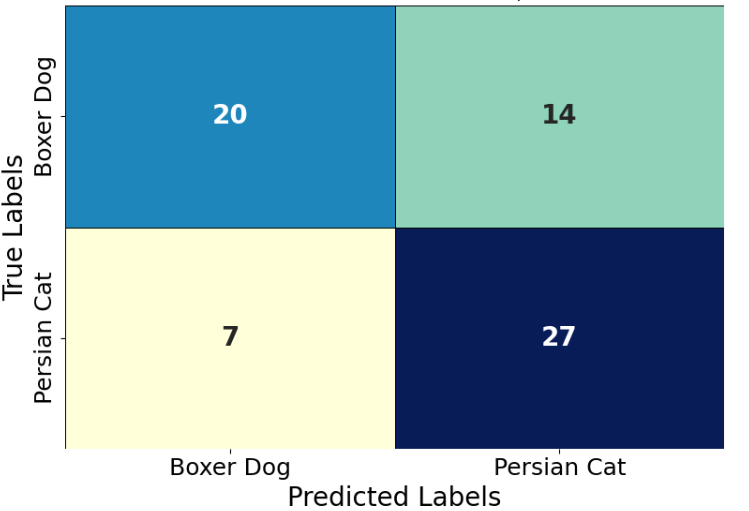}
    \caption{Confusion matrix of the test set for frobenius norm and rank = 10.}
    \label{fig:confusion}
\end{figure}

\begin{figure}[H]
    \centering
    \includegraphics[width=1\linewidth]{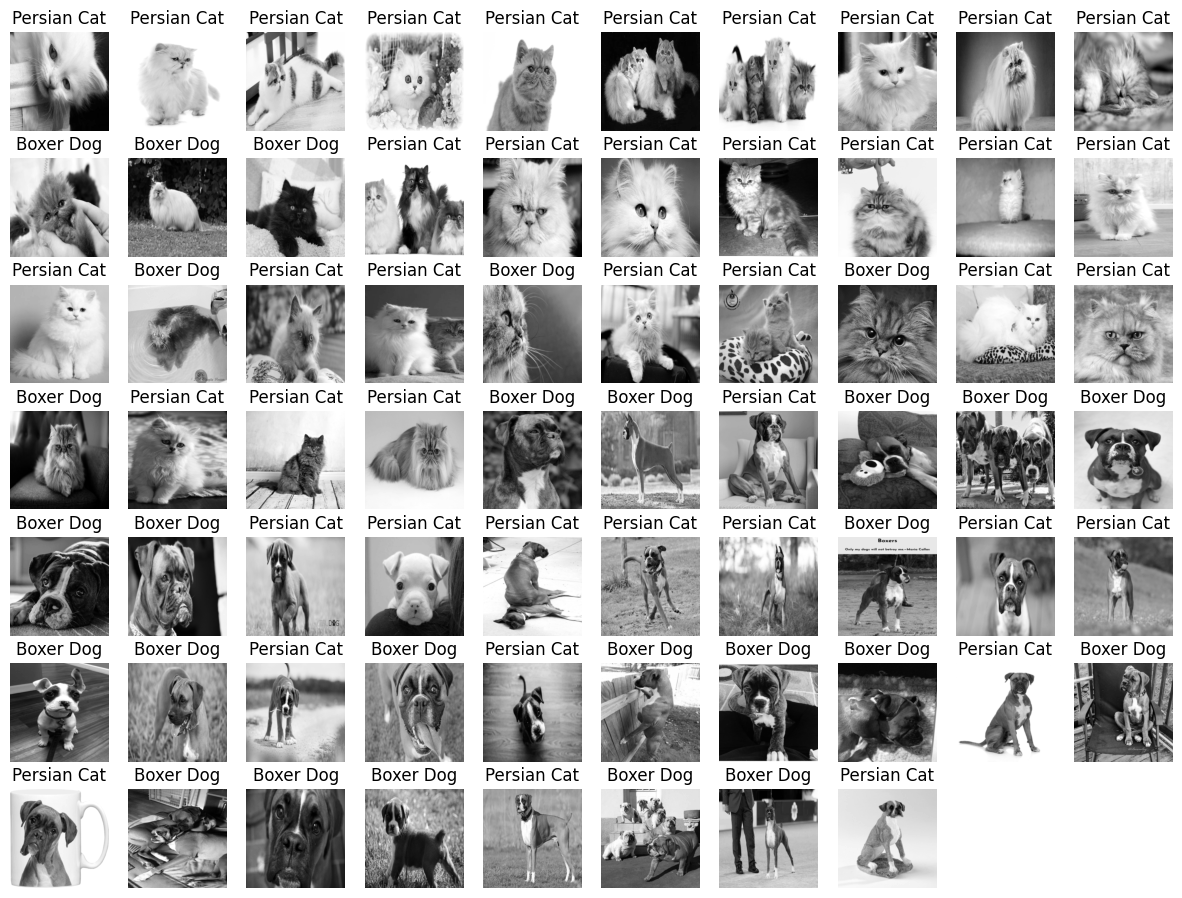}
    \caption{Annotated set of test images using the predicted labels.}
    \label{fig:annotated}
\end{figure}

Figure \ref{fig:confusion} presents the confusion matrix of the test set using the optimal norm and optimal rank computed from the training set. This quantifies the exact number of images from the test set predicted correctly in each class and those predicted incorrectly. From these results, various classification metrics can be calculated. The overall model accuracy on testing set is \textbf{69\%}. The test images are annotated with the predicted class labels using the proposed method, as shown in Figure \ref{fig:annotated}.

\section{Discussion}

This study focuses on specific breeds of cats and dogs distinguished primarily by fur color. The results validate the capability of SVD to effectively capture dominant features through low-rank approximations.

The approach relies heavily on a single feature, fur color. This may not generalize to datasets where the distinguishing characteristics are more complex or subtle. The method's performance is also influenced by the uniformity of the dataset's background. Variations in background color or texture could interfere with the classification accuracy, particularly when grayscale images are used, as background noise may dominate low-rank approximations. Employing images with neutral or consistent backgrounds could help mitigate this limitation.

One potential direction for improving classification performance is to utilize all the RGB channels instead of converting images to grayscale which might discard valuable color information. By incorporating all three color channels, the method could capture more nuanced variations in color intensity and texture, leading to more accurate templates and improved classification performance. Future work could explore how SVD and template optimization perform when applied directly to RGB matrices or by treating each channel independently.

Another potential direction for future work is to investigate the impact of the normalization step during image preprocessing on classification performance. This can be evaluated using the \verb"PIL" library in Python, which does not apply default normalization.

\section{Conclusion}

The image classification results of the proposed method highlight both its strengths and limitations. The findings validate that fur color, as a dominant feature, can be effectively captured through low-rank approximations. However, the moderate accuracy reflects the challenges of relying only on a single feature. This suggest that the method’s applicability may be limited to classes which have visually distinct features even at low rank. This study highlights the potential of an SVD-based method for image classification, making it a viable alternative to machine learning in environments with limited computational resources. Future efforts should explore incorporating additional features and applying the method to more complex datasets to improve classification performance and generalization of the method.

\bibliographystyle{icml2023}

\end{document}